%% file: ms.tex
\documentclass[letterpaper, 10 pt, journal, twoside]{IEEEtran}
\usepackage[utf8]{inputenc}
\input{packages}

\input{commands}

\usepackage{etoolbox}
\newbool{review}
\ifbool{review}{
\newcommand{\changed}[1]{{\color{red}#1}}
\newcommand{\removed}[1]{{\color{red}\ifmmode\text{\sout{\ensuremath{#1}}}\else\sout{#1}\fi}}
}{
\newcommand{\changed}[1]{#1}
\newcommand{\removed}[1]{}
}

\title{Low Frequency Sampling in Model Predictive Path Integral Control}

\date{\today}

\begin{document}
\ifbool{review}{
}{}

\author{Bogdan Vlahov$^1$,
Jason Gibson$^1$,
David D. Fan$^2$,
Patrick Spieler$^2$,\\
Ali-akbar Agha-mohammadi$^{2}$,
Evangelos A. Theodorou$^1$
\thanks{Manuscript received: October, 12, 2023; Accepted March, 11, 2024.}
\thanks{This work was supported in part by the Jet Propulsion Laboratory, California Institute of Technology, under contract with the National Aeronautics and Space Administration (80NM0018D0004) and the Defense Advanced Research Projects Agency, and in part by the Office of Naval Research (N00014-21-1-2074). (\textit{Corresponding author: Bogdan Vlahov})}
\thanks{$^{1}$ Bogdan Vlahov, Jason Gibson, and Evangelos Theodorou are with  the Autonomous Control and Decision Systems Lab, Georgia Institute of Technology, Atlanta GA 30313 USA (e-mail: bvlahov3@gatech.edu; jgibson37@gatech.edu; evangelos.theodorou@gatech.edu)}
\thanks{$^{2}$ David D. Fan, Patrick Spieler, and Ali-akbar Agha-mohammadi are with NASA Jet Propulsion Laboratory, California Institute of Technology, Pasadena, CA, USA (e-mail: david.fan@gmail.com; patrick.spieler@jpl.nasa.gov; aliagha4@gmail.com)}
}

\maketitle
\begin{abstract}
Sampling-based model-predictive controllers have become a powerful optimization tool for planning and control problems in various challenging environments. In this paper, we show how the default choice of uncorrelated Gaussian distributions can be improved upon with the use of a colored noise distribution. Our choice of distribution allows for the emphasis on low frequency control signals, which can result in smoother and more exploratory samples. We use this frequency-based sampling distribution with Model Predictive Path Integral (MPPI) in both hardware and simulation experiments to show better or equal performance on systems with various speeds of input response.
\end{abstract}


\acresetall{}
\section{Introduction}
\label{sec:intro}

\IEEEPARstart{A}{s} autonomous \changed{systems} grow in interest, the choice of methods and algorithms used to do real-time motion planning and control becomes critical to achieve complex tasks.
In general, there are two approaches to this sort of problem: gradient-based and sampling-based.
Gradient-based approaches, such as \ac{DDP} \cite{jacobson1970differential} or \ac{SQP} \cite{boggs1995sequential}, generally have requirements on the dynamics or cost functions used such as they need to be continuously differentiable.
In exchange, they provide controls that converge on the true optimal sequence as the number of iterations increase.
Sampling-based methods such as \cite{rubinstein1999cross,williams2016aggressive} relax the requirements on the dynamics and cost functions to allow for completely arbitrary functions but require many samples to get a good estimation of the true optimal control trajectory.

\ac{MPPI} is one such sampling-based algorithm that has been used to achieve aggressive behavior in a small-vehicle setting. While ideally, one would sample every possible control trajectory to determine the true path integral, this suffers from the curse of dimensionality as both the control dimensions and the time horizon increase. By instead sampling from a Gaussian distribution, computationally-feasible solutions has been derived in path-integral \cite{williams2016aggressive}, information-theoretic \cite{williams2018information}, and stochastic search \cite{wang2021adaptive} approaches. The Gaussian distribution gives a clean equation for distribution calculations and is a natural starting point when constructing a sampling-based algorithm.

However, sampling a Gaussian at every time step leads to control trajectories samples containing high-frequency noise.
Depending on the dynamics model, the effect of the high frequency controls can be dampened when calculating the state trajectory, allowing for control trajectories with high-frequency noise and low-frequency noise to produce similar costs.
This in turn allows the approximate optimal control trajectories computed from finite samples to chatter significantly.
This can cause damage over time when applied to real systems. When looking at data collected from human experts on these systems, it can be seen that they generally take smoother actions over a longer time horizon to achieve their desired behavior.

\begin{figure}[t!]
    \centering
    \includegraphics[width=0.8\linewidth]{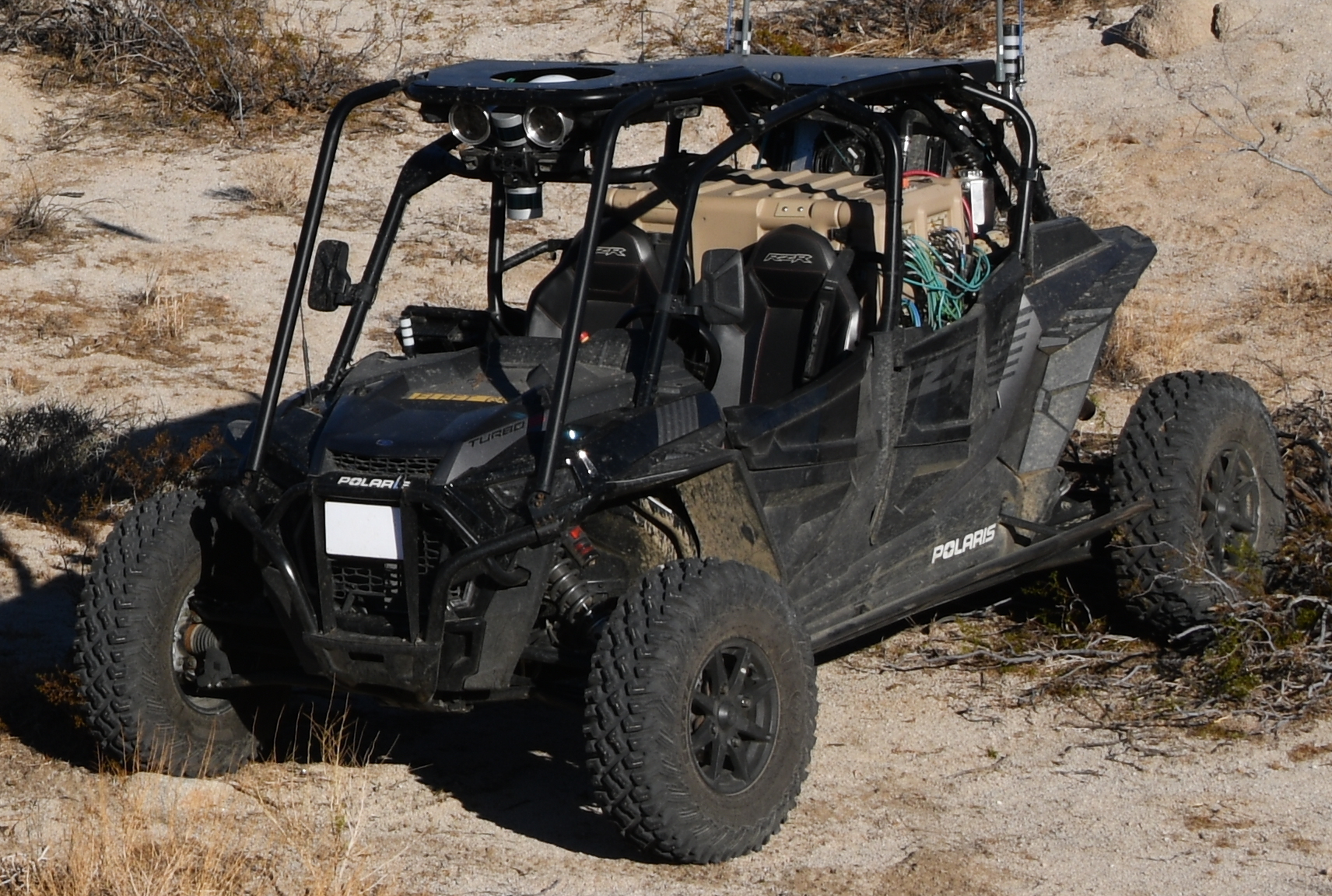}

    \includegraphics[width=0.8\linewidth]{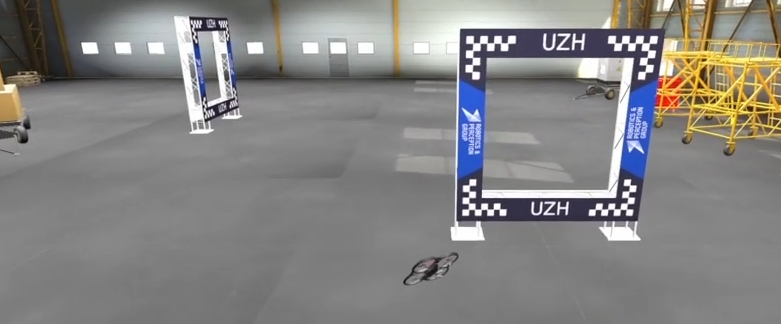}
    \caption{\changed{\textbf{(Upper)} The off-road vehicle in a desert terrain just before an autonomy test. \textbf{(Lower)} A screenshot of the Flightmare quadrotor simulator}}
    \label{fig:hw_vehicle}
\end{figure}

\begin{figure}[t!]
    \centering
    \includegraphics[width=0.88\linewidth]{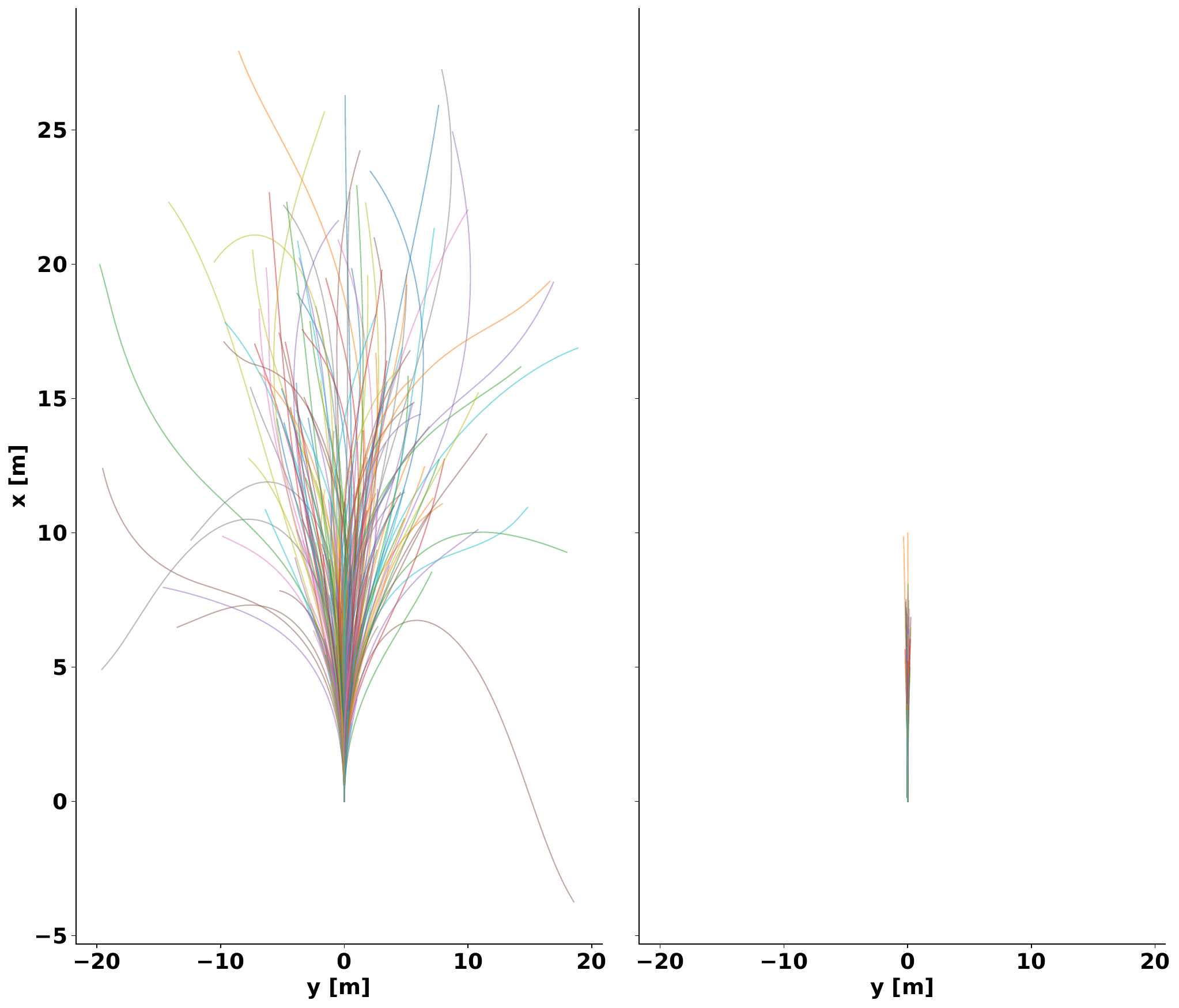}

    \includegraphics[width=0.72\linewidth,clip,trim=5 160 15 10]{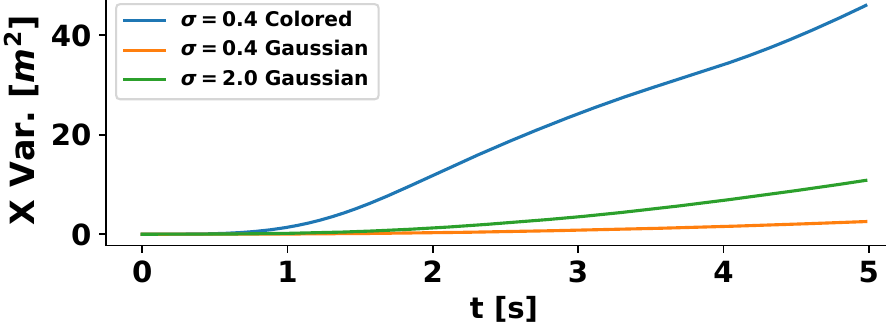}
    \caption{\changed{300 samples of state trajectories generated from running frequency-based \textbf{(Left)} and Gaussian \textbf{(Right)} control samples through the off-road vehicle dynamics described in \cref{subsec:hw_platform}. The frequency-based throttle and steering samples are generated using $\gamma = 4$ and both distributions used the same $\sigma$ set to $0.4$. Descriptions of $\gamma$ and $\sigma$ can be found in \cref{subsec:colored_noise_def}. \textbf{(Lower)} Variance in the x direction of the same 300 samples from frequency-based and Gaussian distributions over time. There is an additional Gaussian distribution with $\sigma = 2.0$ to show that increasing the standard deviation does not allow Gaussian sampling to explore as far as colored samples. Colored samples reach further extremes in both the $x$ and $y$ directions due to the reduction of high frequency signals in the control trajectories.}}
    \label{fig:motivating example}
\end{figure}

In order to address these issues, we look to sampling distributions that can adjust the level of high-frequency noise that appears in control trajectories.
In this paper, we explore sampling focused around the low-frequency domain by using a colored noise distribution.
We show how this choice of sampling distribution can lead to better state exploration when compared to standard Gaussian sampling in \ac{MPPI}, like shown in \cref{fig:motivating example}.
Furthermore, we will show the resulting controls are smoother and thus reduce the wear and tear on \removed{real} systems, such as \changed{those shown in} \cref{fig:hw_vehicle}, caused by chattering.
\changed{Previous works have attempted to sample smoother controls in various ways but our method provides an additional parameter, $\gamma$, to adjust the smoothness of sampled control without adjusting the overall range of sampled controls like adjusting variance would cause.
This smoothness adjustment is done by explicitly lowering the chance of sampling of higher frequency signals, resulting in less oscillatory behavior and allows its use with multiple systems with different control bandwidths.}
The contributions of this paper can be summarized as follows:
\begin{itemize}
    \item We show how a frequency-based sampling distribution can be used in \ac{MPPI} with minimal adjustments to the update rules and optimal control calculation in \cref{sec:math}.
    \item We perform experiments on a real hardware platform that has large control lag in \cref{subsec:hw_platform} as well as highly-reactive systems in \cref{subsec:sim_quad_results,subsec:double_integrator}. These tests show that our approach can be considered a generalization of Gaussian sampling applicable in many scenarios.
\end{itemize}

The paper is organized as follows. In \cref{sec:related_work}, we go over other approaches to choosing sampling distributions.
In \cref{sec:math}, we discuss the frequency-based sampling is performed and the modifications required to \ac{MPPI}. We showcase experiments in \cref{sec:results} and conclude in \cref{sec:conclusion}.

\section{Related Work}
\label{sec:related_work}
\changed{Before going into frequency-based techniques, a brief overview of relevant concepts is required. The \ac{PSD} provides a measure of how much power every frequency, $f$, contributes to a time-domain signal. Samples from a zero-mean, uncorrelated Gaussian distribution have constant power at every frequency, $PSD(f) \propto 1$. Colored Noise distributions cover a large space of distributions that have \iac{PSD} of the form $PSD(f) \propto \frac{1}{f^\gamma}$.}

There have been multiple works on adjusting the sampling distributions used in \ac{MPPI} to improve sampling exploration.
The original Gaussian sampling distribution used in \cite{williams2016aggressive} provides a good starting point, but can have difficulty to fully explore the possible state space.
\cite{watson2023inferring} uses \acp{GP} as their sampling distribution and can show that the resulting optimal controls are smoother than those found using Gaussian noise.
However, it could require much more computation to generate samples depending on the time horizons chosen as the prior.
\cite{mohamed2022autonomous} uses \changed{\iac{NLN}} distribution for sampling and shows that the samples generated can explore a wider space than standard Gaussian sampling. While \changed{\ac{NLN}} distributions allow for more sampling on the tail ends of a distribution compared to a normal distribution, there is still only a small chance to sample multiple tail-end values in a row for slow-acting systems unlike colored noise distributions.
\changed{\cite{kim2022smooth} showed that sampling in a derivative "action space" would also further improve the exploration and smoothness of the control trajectories \ac{MPPI} would produce.
Specifically, the authors end up not integrating the Gaussian samples over time but instead over iterations of \ac{MPPI}.
While this sampling technique does improve smoothness, it does so by multiplying the samples by $dt$. This can cause explorations issues with small $dt$.
Our frequency-based sampling can adjust the time correlation of our trajectories independent of $dt$.}

Other works have used multi-hypothesis distributions \cite{wang2021variational,lambert2021stein} such as \acp{GMM} and Stein Variational distributions to try to increase the exploration capabilities of the algorithm.
While \acp{GMM} can maintain multiple distributions, there is nothing to prevent the several modes from collapsing which became the motivation for Stein variational policies.
Stein variational policies maintain multiple particles as Gaussian distributions with fixed variance but ensure that the means of these Gaussians are pushed apart by inclusion of a kernel function. Recently, there has also been work in learning sampling distributions with concepts like normalizing flow \cite{sacks2023learning,power2022variational}. These methods make use of a starting distribution and then perform a transform of that base distribution into a new one. This transform can then make use of machine learning to incorporate outside information such as the location of obstacles in order to bias the resulting distribution into safe regions.

The use of frequency-based sampling has been explored before as well.
\cite{pinneri2021sampleefficient} used frequency-based sampling as well as other techniques in a \ac{CEM} controller to create \ac{iCEM}.
They showed that using frequency-based sampling has one of the largest positive impacts on the overall performance of the \ac{CEM} controller.
We will use the same frequency-based sampling technique with \ac{MPPI} which uses a weighted combination of every sample to compute the optimal control trajectory rather than using a percentage cutoff.
\cite{zhang2021asynchronous} used a power law noise to improve exploration for their \ac{RL} agent.
They specifically generated a $\frac{1}{f^2}$ distribution by filtering white noise and saw large improvements over Gaussian sampling in their ablation study.
We will be using a different power law noise generation technique so that we can use any exponent we desire.

\begin{figure}[ht!]
    \centering
    \includegraphics[width=0.95\linewidth,clip,trim=0 220 0 0]{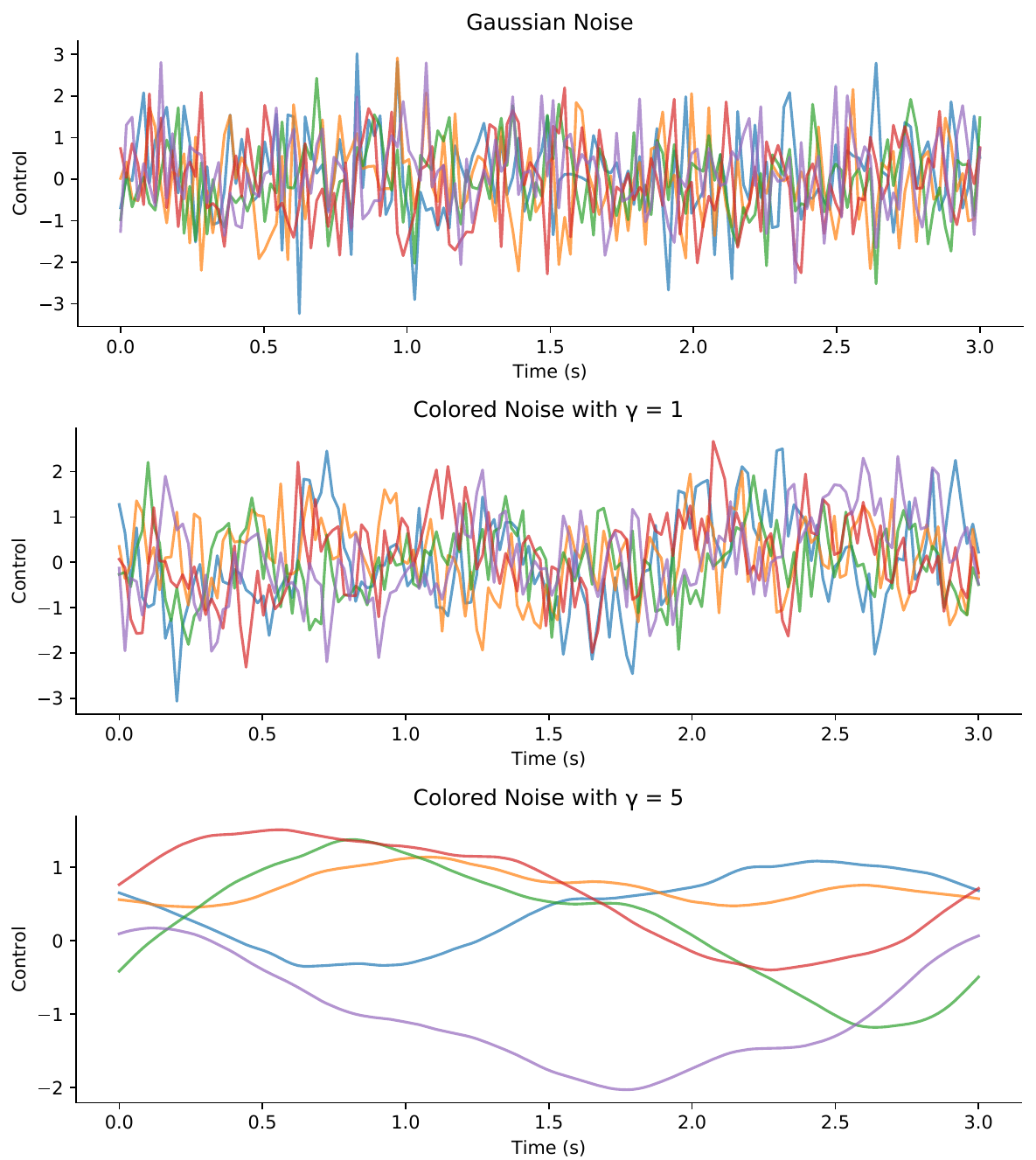}
    \caption{A comparison of Gaussian and Colored noise with various exponents. As $\gamma$ increases, the samples generated become smoother and can reach tail-end values more consistently.}
    \label{fig:colored_sampling}
\end{figure}
\section{Mathematical Background}
\label{sec:math}
\subsection{Low Frequency Sampling Distribution}
\label{subsec:colored_noise_def}
\removed{The \ac{PSD} provides a measure of how much power every frequency, $f$, contributes to a time-domain signal. Samples from a zero-mean, uncorrelated Gaussian distribution have constant power at every frequency, $PSD(f) \propto 1$. Colored Noise distributions cover a large space of distributions that have \iac{PSD} of the form $PSD(f) \propto \frac{1}{f^\gamma}$.}

For this paper, we will be using a colored noise distribution found by sampling Gaussians in the frequency domain and performing the  \ac{iFFT} to go to the time domain, first described in \cite{timmer1995generating}. These samples are control trajectories with a finite time horizon of $T$ steps. We first will show that samples derived from this method are still Gaussian in the time domain.

In order to limit our time-domain signal to contain only real components, we sample a particular frequency-domain sequence known as a \changed{Hermitian-symmetric sequence \cite[Fig. 2.5]{bracewell1966fourier}}. A \changed{Hermitian-symmetric} sequence is one where the second half of the sequence is the complex conjugate of the first half of the sequence.
Let $Z[n] = Z_{real}[n] + iZ_{imag}[n]$ be a point in the frequency domain with a real and imaginary component.
\changed{There will be $N= \lfloor \frac{T}{2} \rfloor + 1$ frequency samples.
The sampling of $Z_{real}[n]$ and $Z_{imag}[n]$ is done from Gaussian distributions of the form $\mathcal{N}\left(\mu_n,\left(\max\left\{\frac{n}{N},f_{min}\right\}\right)^{-\gamma}\frac{\sigma^2}{\zeta}\right)$ where $f_{min} = \frac{1}{N}$ is a cutoff frequency to ensure non-zero variance, $\gamma$ is a user-chosen colored exponent, $\sigma$ is a user-chosen standard deviation, and
\begin{align}
\zeta = T^{-2}N^\gamma\left(1 + 4\sum_{n=1}^{N-1} n^{-\gamma} \right)
\end{align}
ensures that the time-domain variance equals $\sigma^2$.
The effects of different $\gamma$ on sample control trajectories is shown in \cref{fig:colored_sampling}.
The Hermitian-symmetric} sequence can be compactly represented as
$Z_N = \left\{Z[0], Z[1], ..., Z[N-1] \right\}$
in the frequency domain \removed{with $N= \lfloor \frac{T}{2} \rfloor + 1$ sample points instead of $T$}.
As each point $Z[n]$ in the frequency domain has both a real and imaginary part, there is no reduction of the total number of points sampled compared to uncorrelated time-domain Gaussian sampling. The full frequency-domain sequence, $Z^\prime_T$ can then be constructed as
\begin{align}
    Z^\prime[t] = \begin{cases}
        Z[t] & \text{if}\ 0 \leq t \leq \changed{N-1} \\\\
        \overline{Z[T-t]} & \text{if}\ \changed{N-1} < t < T
    \end{cases},
\end{align}
where $\overline{\cdot}$ is the conjugate operator.
The time-domain sequence is found using the \ac{iFFT}, referred to as $\Psi$ for notional convenience:
\begin{align}
z(t) &= \frac{1}{T} \sum_{n=0}^{T-1} Z^{\prime}[n] e^{i2\pi  nt/T} = \Psi\left(Z_N\right).
\end{align}

The \ac{iFFT} is thus a matrix transform of our frequency-domain samples and can be simplified to the following when \changed{sampling Hermitian-symmetric sequences:
\begin{align}
&\begin{bmatrix}
z(0) \\
\vdots \\
z(t) \\
\vdots \\
\end{bmatrix} = M \begin{bmatrix}
Z_{real}[0] \\
\vdots \\
Z_{real}[n] \\
Z_{imag}[n] \\
\vdots \\
\end{bmatrix}, \ \forall
\begin{matrix}
    &n = 0, ..., N-1 \\
    &t = 0,..,T-1
\end{matrix}
\label{eq:ifft_matrix_transform}
\end{align}
\begin{align}
M =\frac{1}{T}
\begin{bmatrix}
1 
& \cdots & 2 &  0 & \cdots \\
& & \vdots& &  \\
1 
& \cdots & 2\cos\left(\frac{2\pi nt}{T}\right) &  -2\sin\left(\frac{2\pi nt}{T}\right) & \cdots \\
& & \vdots& &
\end{bmatrix}.
\end{align}}
Note that if $T$ is even, the final sample, $Z[\changed{N-1}]$, must only have a real component $\left(Z_{imag}[\changed{N-1}] = 0\right)$ to be considered Hermitian. As the $M$ matrix does not depend on the samples drawn for $Z[n]$, each $z(t)$ is a linear combination of Gaussian independent random variables, making it Gaussian as well. However, these samples will now be time-correlated. In a slight abuse of notation, we will use the term Gaussian to mean uncorrelated Gaussian distributions and Colored to refer to our frequency-based sampling technique throughout the rest of this paper unless otherwise noted.

\removed{The sampling of $Z_{real}[n]$ and $Z_{imag}[n]$ is done from Gaussian distributions of the form $\mathcal{N}\left(\mu_n,\left(\max\left\{\frac{n}{N},f_{min}\right\}\right)^{\gamma}\sigma^2\right)$ where $f_{min} = \frac{1}{N}$ is a cutoff frequency to ensure non-zero variance and $\sigma$ is a user-chosen standard deviation.}

\subsection{MPPI derivation from Frequency Domain Sampling}
\label{subsec:mppi_derivation}
Consider a general nonlinear system with discrete-time dynamics and cost function of the following form:
\begin{align}
    \vb{x}_{t+1} &= \vb{x}_t + \vb{F}\left(\vb{x}_t, \vb{u}_t\right) \Delta t + \vb{w}_t, \\
    \vb{J}(X, U) &= \phi(\vb{x}_T) + \sum_{t = 0}^{T -1}q(\vb{x}_t),
\end{align}
where $\vb{x} \in \R^{n_x}$ is our state, $\vb{u} \in \R^{n_u}$ is our control, and $\vb{w} \in \R^{n_x}$ are external disturbances that are unknown but assumed to be bounded. $\vb{J}(\cdot, \cdot)$ is the cost function of a given state trajectory $X$, and control trajectory $U$, with a terminal cost $\phi(\cdot)$ and nonlinear state cost $q(\cdot)$. \changed{We exclude control costs from this work for the sake of brevity but they can be added with no change to the following discussion.}

\ac{MPPI} is trying to minimize the cost function of the system, by sampling and applying a weighted exponential averaging to produce the optimal control.
By following the stochastic search derivation of \ac{MPPI}, we can make some small adjustments to accommodate our new sampling distribution.
\changed{We make use of the fact that Gaussian distributions belong to the exponential family of probability distributions $P$, parameterized by $\theta$, with $\theta = [\mu, \Sigma]^T$ specifically for Gaussians.
We define $l(\theta)$ in order to better align with theory presented in \cite{wang2021adaptive},
\begin{align}
    l(\theta) &= \ln{\Expectation[\epsilon \sim P(\theta)]{\Shape(-\J(X,U))}}.
\end{align}
where $\Shape(\cdot)$ is a generic shaping function.
This function $l(\theta)$ represents a log-transform on the expectation of the negated cost function over sampled controls. Maximizing $l(\theta)$ is equivalent to minimizing $\Expectation[]{\J(X,U)}$ which is the goal of our original optimal control problem.
\cite{wang2021adaptive} shows that, given $l(\theta)$, the update law for $\theta$ at iteration $k$ can be gradient ascent,
\begin{align}
    \theta^{k+1}_{t} &= \theta^{k}_{t} + \alpha_{k} \nabla_{\theta_{t}}l(\theta^{k}_{t}) \label{eq:grad_ascent_law} \\
    \nabla_{\theta_{t}}l(\theta_{t}) &= \frac{\Expectation[\epsilon_t \sim P(\theta)]{\Shape(-\J(X,U))\nabla_{\theta_{t}}\ln{p\left(\epsilon_t; \theta_t\right)}}}{\Expectation[P(\theta)]{\Shape(-\J(X,U))}}, \label{eq:loss_gradient_wrt_theta}
\end{align}
where $\alpha_{k}$} satisfies:
\begin{align}
    \alpha_{k} > 0 \quad\forall k, \ \lim_{k \rightarrow \infty} \alpha_{k} = 0, \ \sum_{k = 0}^{\infty} \alpha_{k} = \infty,
    \label{eq:step_size_assumptions}
\end{align}
in order to achieve asymptotic convergence \cite{zhou2014gradientbased}.
If the shaping function is chosen to be the exponential function,
\begin{align}
    \Shape(y;\lambda) = \expf{\frac{1}{\lambda} y},
\end{align}
with $\lambda$ being the inverse temperature, \changed{\cref{eq:loss_gradient_wrt_theta} becomes equivalent to the update law for information-theoretic \ac{MPPI} \cite{williams2018information},
\begin{align}
    \vb{u}^*_t &= \Expectation[V \sim P(\theta)]{w(V)(\vb{v}_t)},
    \label{eq:information_theoretic_mppi_update}\\
    w(V) &= \frac{1}{\eta}\expf{-\frac{1}{\lambda}\J\PP{X,V}}, \label{eq:mppi_weight}
\end{align}
where $\eta$ is a regularization term, $V=\{\vb{v}_0, ..., \vb{v}_{T-1}\}$ is a sampled control trajectory from the Gaussian control distribution $P(\theta)$, and $w$ is the weight associated with $V$.}
As our distribution in the frequency domain is independent and Gaussian, we can change the $t$ variable in \cref{eq:grad_ascent_law,eq:loss_gradient_wrt_theta} to $n$ and subsume the \ac{iFFT} transform inside of $\J$ with the following equation, $\Jf(X, U) = \J(X, \Psi(U))$. This allows us to use the same update law to update the parameters of our frequency-domain sampling distribution,
\begin{align}
    \theta^{k+1}_{n} &= \theta^{k}_{n} + \changed{\alpha_{k}} \nabla_{\theta_{n}}l(\theta^{k}_{n}), \label{eq:grad_ascent_law_freq} \\
    \nabla_{\theta_{n}}l(\theta_{n}) &= \frac{\Expectation[\changed{\epsilon}_n \sim P(\theta)]{\Shape(-\Jf(X,U))\nabla_{\theta_{n}}\ln{p\left(\changed{\epsilon}_n; \theta_n\right)}}}{\Expectation[P(\theta)]{\Shape(-\Jf(X,U))}}. \label{eq:loss_gradient_wrt_theta_freq}
\end{align}
For this paper, we choose to only update the mean of our control distribution, $\theta_n = \{\mu_n\}$, and leave the variance constant; variance updates would be derived the same way if desired.
The gradient is estimated using Monte-Carlo sampling \removed{as in \cite{wang2021adaptive}}.
Once the gradient is estimated, we choose to do the mean trajectory update in the time domain for computational convenience,
\begin{align}
\mu_t^{k+1} = \Psi\left(\mu_N^{k+1}\right) = \Psi\left(\mu_N^{k}\right) +
\alpha_{k}\Psi\left( \nabla_{\mu_{N}}l\left(\mu_N^{k}\right)\right),
\end{align}
where $\Psi\PP{\mu_N^k}$ is equivalent to $\mu_t^{k}$ and
\begin{align}
\Psi\PP{\nabla_{\mu_N}l\PP{\mu_N^k}} = \Psi\PP{\sum_{\changed{m}=0}^M w_\changed{m} \mathcal{E}_N^\changed{m}} = \sum_{\changed{m}=0}^M w_\changed{m} z^\changed{m}(t),
\end{align}
with $M$ being the total number of sample trajectories, $\mu_N$ referring to the collection of $\{\mu_n; \ n = 0,..,N\}$, \changed{$\mathcal{E}^m_N$ refers to the $m$th sample trajectory's collection of frequency samples $\mathcal{E}^m_n = \{\epsilon^m_0, \epsilon^m_1, ..., \epsilon^m_N\}$, and $w_m$ is calculated using \cref{eq:mppi_weight}}.
The final algorithm for Frequency Sampling \ac{MPPI} can be seen in \cref{alg:frequencyMPPI} with changes from the standard \ac{MPPI} algorithm highlighted in blue.
\begin{algorithm}[ht]
\footnotesize
\LinesNumbered
\SetKwInOut{Input}{Given}
\Input{
    $\vb{F}\left(\cdot, \cdot\right)$, $q\left(\cdot\right)$, $\phi\left(\cdot\right)$ $M$, \changed{$I$}, $T$, $\lambda$, $\sigma$, $\mathbb{\gamma}, \changed{\zeta,} f_{min}, \alpha$: System dynamics, running state cost, terminal cost, num. samples, \changed{num. iterations}, time horizon, temperature, standard deviations, sampling exponents\changed{, sampling normalization term}, minimum sampling frequency, update step size\;
}
\SetKwInOut{Input}{Input}
\Input{
    $\vb{x}_0$, $\vb{U}$: initial state, mean control sequence\;
      }
\SetKwInOut{Input}{Output}
\Input{
    $\mathcal{U}$: optimal control sequence
}
\BlankLine

\tcp{Calculate Frequency range}
$N \leftarrow \lfloor \frac{T}{2} \rfloor + 1$\;
\tcp{Begin Cost sampling}
\For{\changed{$i \leftarrow 1$ \KwTo $I$}}{
\For{$m \leftarrow 1$ \KwTo $M$}{
    $J^m \leftarrow 0$\;
    $\vb{x} \leftarrow \vb{x}_0$\;
    {\color{blue}\tcp{Sample Frequency noise}
    $\mathcal{E}^m_N = \left( \epsilon_0^m \dots \epsilon_{N}^m \right), ~\epsilon_n^m \in \mathcal{N}(0, \changed{\zeta^{-1}}\left(\max\left\{\frac{n}{N},f_{min}\right\}\right)^{\changed{-\gamma}})$\;}
    \For{$t \leftarrow 0$ \KwTo $T-1$}{
        {\color{blue}$z^m(t) \leftarrow \Psi\left(\mathcal{E}_N^m\right)$\;}
        $\vb{v}_{t} \leftarrow \vb{u_t} + \sigma z^m(t)$\;
        $\vb{x} \leftarrow \vb{F}\left(\vb{x}, \vb{v}_t\right)$\;
        $J^m \pe q\left(\vb{x}\right)$;
    }
    $J^m \pe \phi\left(\vb{x}\right)$
}
\tcp{Compute trajectory weights}
$\rho \leftarrow \min\left\{J^1, J^2, ..., J^\changed{M}\right\}$\;
$\eta \leftarrow \sum_{\changed{m}=1}^{\changed{M}}\expf{-\frac{1}{\lambda}\left(J^\changed{m} - \rho\right)}$\;
\For{$\changed{m} \leftarrow 1$ \KwTo $\changed{M}$}{
    $w_\changed{m} \leftarrow \frac{1}{\eta} \expf{-\frac{1}{\lambda}\left(J^m - \rho\right)}$\;
}

\tcp{Control update}
\For{$t \leftarrow 0$ \KwTo $T-1$}{
    $\mathcal{U}_t \leftarrow \vb{u}_t + \alpha \sum_{m=1}^{M}w_m z^m(t)$\; \label{alg:policy-update-line}
}
}
\caption{Frequency-based Sampling \ac{MPPI}}
\label{alg:frequencyMPPI}
\end{algorithm}
\section{Experimental Results}
\label{sec:results}
\subsection{Off-road Vehicle Platform}
\label{subsec:hw_platform}
We conducted experiments on a full-scale autonomous off-road modified \changed{Polaris Razer X}, shown in \cref{fig:hw_vehicle}.
\changed{It is equipped} with a sensor array mounted on top and a onboard computer using 4 RTX 3080s, 256 \si{\giga\byte} of RAM, and a \changed{Threadripper 3990x}.
The autonomy controls the vehicle using the power-assisted steering wheel actuator, \changed{a pump for brake pressure, and direct access to the throttle}.
The control inputs for the throttle and brake are combined to create a range of $[-1, 1]$ and the steering wheel control input is normalized to within $[-1, 1]$.
Each control input has some form of delay from applying a signal and seeing an effect on the vehicle.
For example, turning the steering wheel fully from right to left takes about two seconds.
We learned the dynamics and modeling delays, $\vb{F}\PP{\cdot, \cdot}$, using \cite{gibson2023multistep}.
The slow response to controls is where the limitations of Gaussian sampling become clear.
In order to achieve tight turning maneuvers like human drivers can, \ac{MPPI} needs to sample large steering values in the same direction for multiple time steps in a row.
\changed{While increasing the variance of the Gaussian distribution makes sampling larger values easier, it does nothing to address the fact that the samples need to have the same sign to actually make a sharp turn.
Running several iterations can move the mean to one extreme but this is computationally expensive and slow-moving.}

In our experiments, both the Gaussian and Colored sampling \ac{MPPI} algorithms \changed{are implemented in CUDA with $I = 3$ iterations, $M = 6144$ samples, $\lambda = 0.1$, $dt = 0.02 \si{\s}$, and $T = 250$} step horizon.
The cost function for both algorithms \changed{is}
\begin{equation}
\begin{split}
    q(\vb{x}) &= \textit{CostToGo}\PP{\vb{x}}
    +\hat{d}\Bigl(\textit{WheelRisk}\PP{\vb{x}}\\
    &+\textbf{BodyRisk}\PP{\vb{x}}
    +\textbf{Lethal}\PP{\vb{x}}
    +\textbf{Rollover}\PP{\vb{x}} \\
    &+\textbf{Roll}\PP{\vb{x}}
    +\textbf{Pitch}\PP{\vb{x}}
    +\textit{Speed}\PP{\vb{x}} \Bigr),
\end{split}
\end{equation}
where $\hat{d} = \vb{v}dt$ is the estimated distance traveled at a given timestep.
\removed{These cost components are further described in the appendix.}

\begin{figure}[hbt!]
    \centering
    \subfloat[Gaussian \label{fig:hw_gaussian_results}]{
        \includegraphics[width=0.5\linewidth,clip,trim=0 0 0 24]{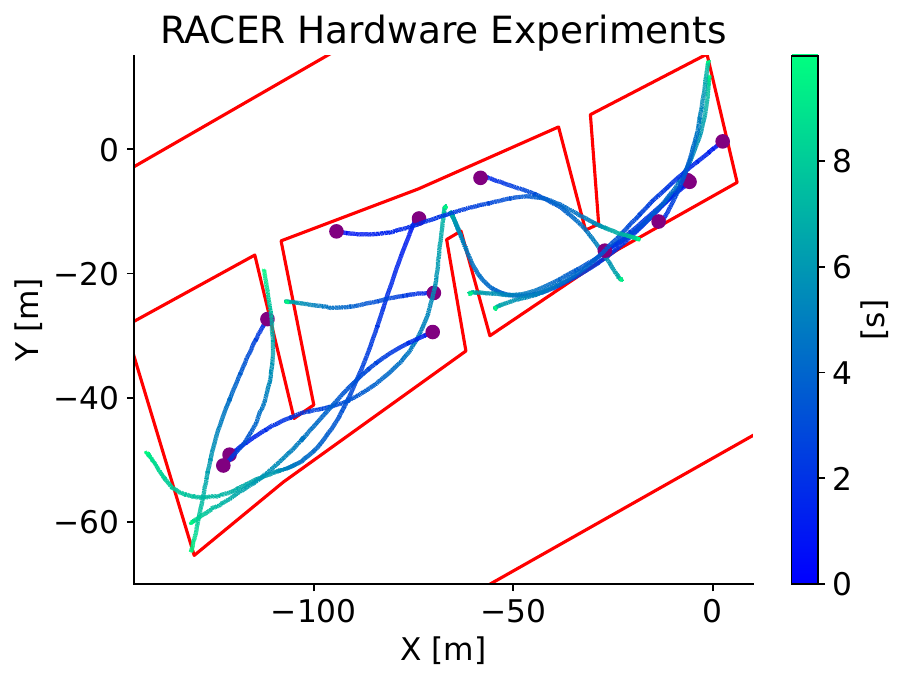}
    }
    \subfloat[Colored
    \label{fig:hw_colored_results}]{
        \includegraphics[width=0.5\linewidth,clip,trim=0 0 0 24]{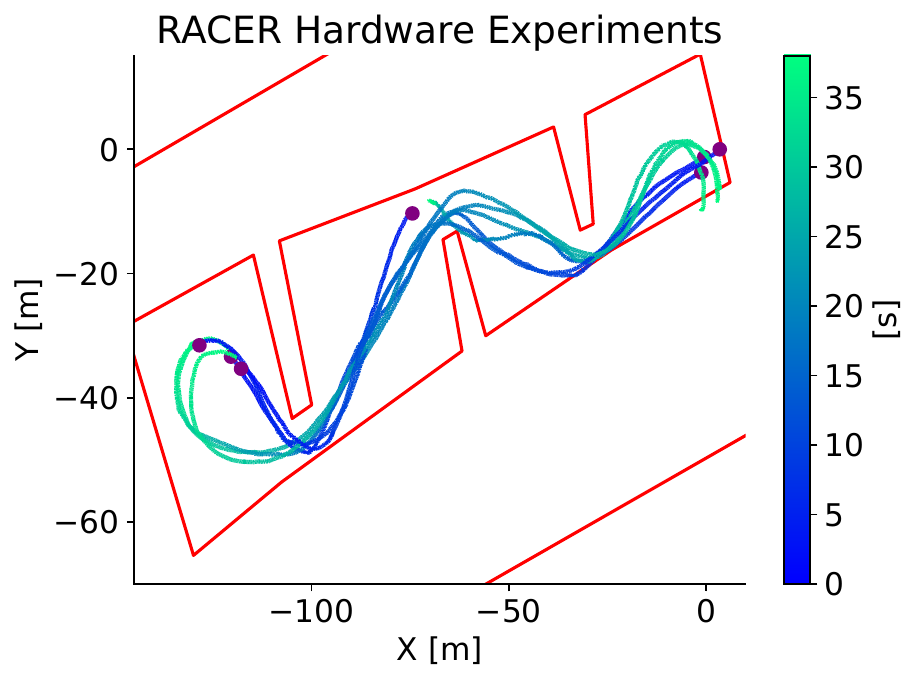}}
    \caption{Hardware experiments using Gaussian and Colored Sampling on the vehicle. These graphs show the vehicle trying to maneuver in a zig-zag corridor shown in red. The colors along the trajectories indicate the amount of time spent in autonomy before manual override/goal achieved \changed{with purple dots indicating the starting points}. The Colored samples can go from one end of the corridor to the other while the Gaussian sampling struggles to make more than a single turn even when started in various locations.}
    \label{fig:hw_results}
\end{figure}

\begin{figure}[hbt!]
    \centering
    \includegraphics[width=0.75\linewidth]{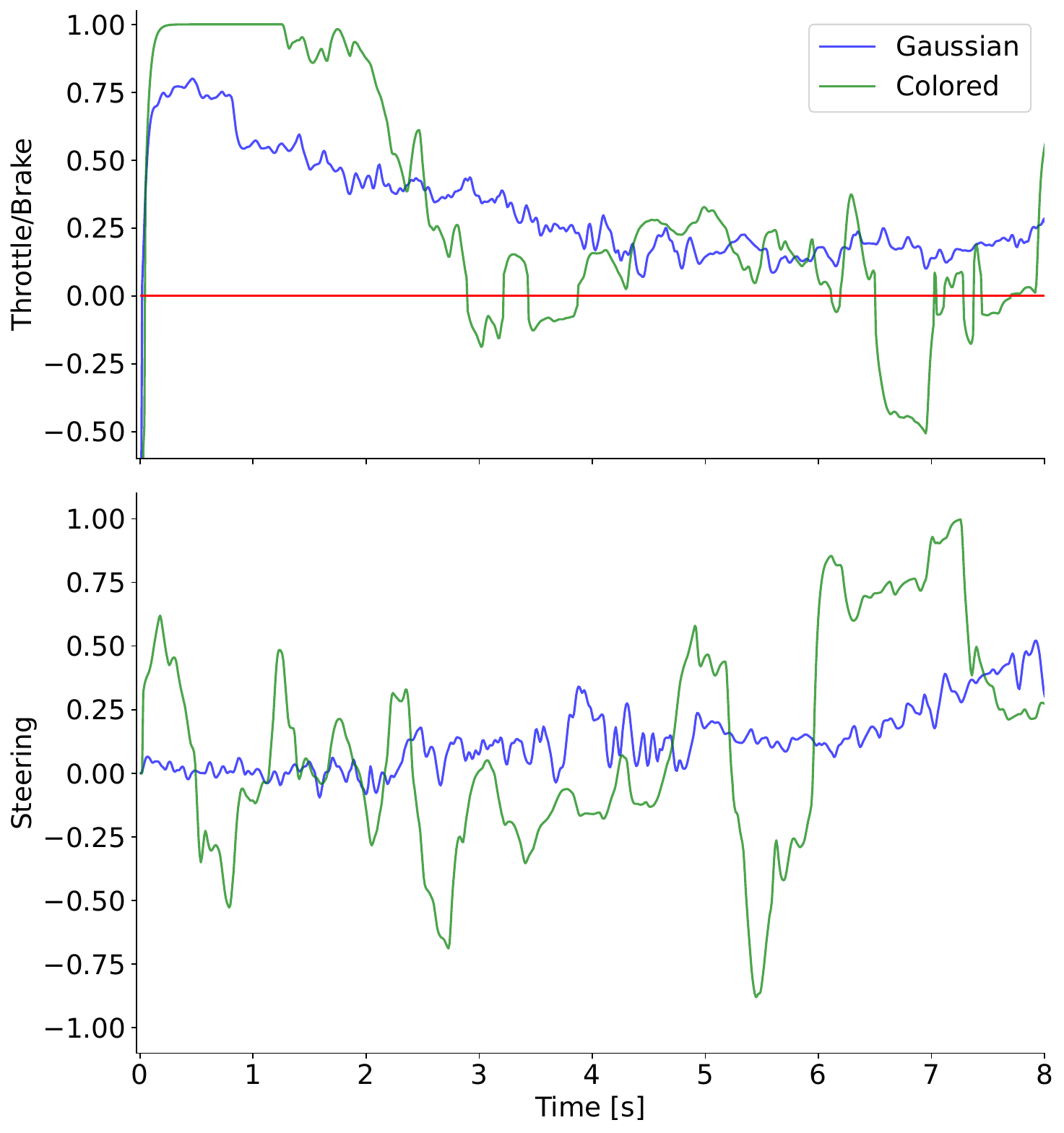}
    \caption{Controls from the vehicle hardware attempts. Picture are the controls from the first eight seconds of the first attempt of each controller starting from the same location. We can see that the colored sampling technique is achieves smoother \changed{and larger} throttle, brake, and steering commands. Throttle and brake are combined into a single graph with brake values being below the red line.}
    \label{fig:fastest_hw_gaussian_control}
\end{figure}

In \cref{fig:hw_results}, the vehicle is attempting to stay within the bounds of this zig-zag corridor which requires hard turns for appreciable periods of time to achieve.
There is a human safety operator in the vehicle that would engage the manual override whenever the vehicle would exit the corridor.
It is important to note that both controllers look to violate the boundaries of the corridor but this is due to drift in state estimation between the vehicle position and the corridor location.
\removed{it can be seen that the} Gaussian sampling could not complete more than a single turn even when started in various locations whereas the \changed{Colored} sampling only had one human intervention in its six attempts.
The times and success rates are summarized in \cref{tab:experiment_summary}.
\changed{The computation times of the Colored \ac{MPPI} averaged $26.846 \si{\milli\s}$ over these experiments while the vanilla MPPI algorithm averaged $24.839 \si{\milli\s}$, showing there is minimal overhead to using the Colored approach on this hardware.}

Furthermore, from the controls plotted in \cref{fig:fastest_hw_gaussian_control}, we can further see that the Gaussian sampler can hit high steering angles of $0.5$ but only slowly as the \ac{MPC} nature of \ac{MPPI} adjusts the mean of the steering trajectory in small steps. For this testing, the control standard deviations were both set to $0.8$, the combined throttle and brake exponent was $\changed{\gamma_{throttle} =} 10$, and the steering exponent was $\changed{\gamma_{steering} =} 4$.

\begin{table}[hb!]
\centering
\caption{Lap Experiment Summaries}
\label{tab:experiment_summary}
\footnotesize

\sisetup{round-mode=places,separate-uncertainty,
detect-weight=true
}
\begin{tabular}{l
S[round-mode=uncertainty,round-precision=3,round-pad=false]
S[round-precision=3]
S[round-precision=2]
}
\toprule
\text{Sampling} & \text{Avg Time[\si{\s}]} & \text{Min Time[\si{\s}]} & $\mathcal{S}_{R}$[\%]   \\ \midrule
Off-road HW Gaussian  & \text{N/A} & \text{N/A} & 0 \\
Off-road HW Colored   & \bfseries \num{38.333 \pm 3.316} & \bfseries 35.99 & \bfseries 83.33333 \\
\midrule
Quad. SIM Gaussian & \bfseries \num{8.546724077333334 \pm 0.2062358000133801} & 8.36005727 & 24.32432432 \\
Quad. SIM Colored  & \num{8.690610210777777 \pm 0.4565232731785332} & \bfseries 7.695057759 & \bfseries 42.85714285 \\
\end{tabular}
\end{table}
\begin{figure*}[ht!]
    \centering
    \subfloat[Gaussian \label{fig:quad_gaussian_results}]{
        \includegraphics[width=0.4\textwidth,clip,trim= 0 0 0 20]{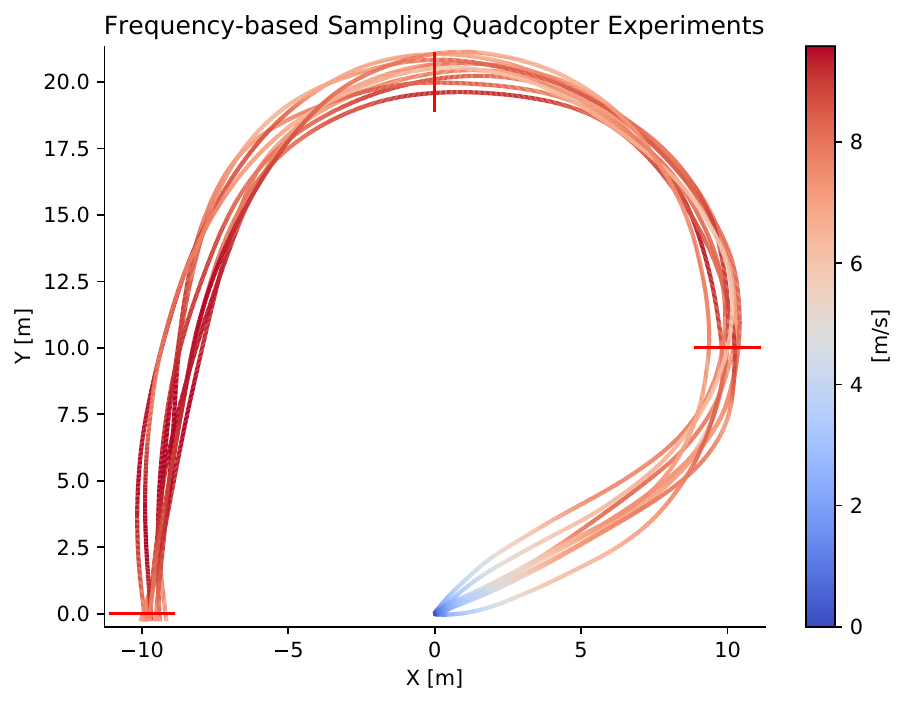}
    }
     \subfloat[Gaussian \label{fig:fastest_quad_gaussian_control}]{
        \includegraphics[width=0.5\textwidth]{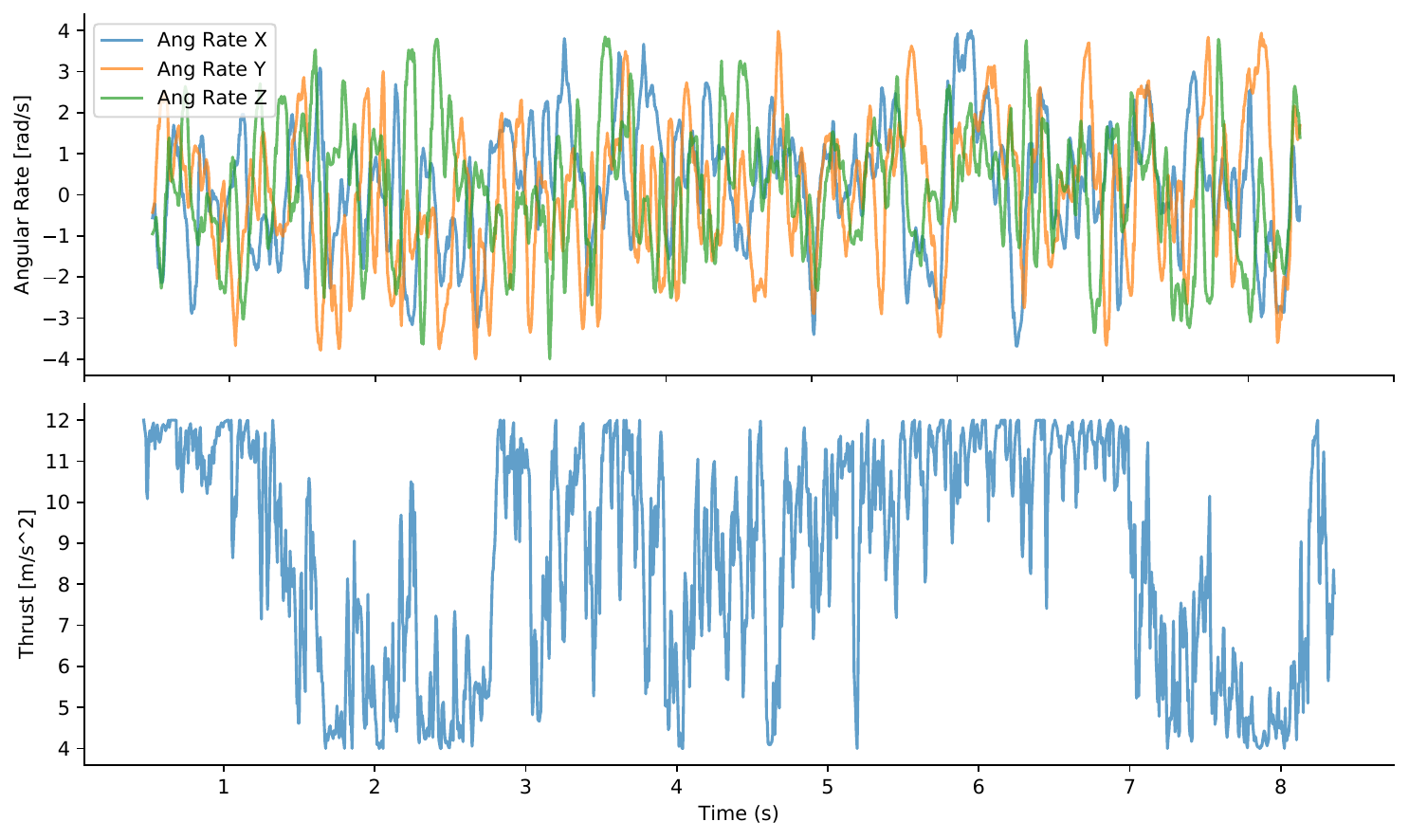}
    }

   \subfloat[Colored \label{fig:quad_colored_results}]{
        \includegraphics[width=0.4\textwidth,clip,trim= 0 0 0 20]{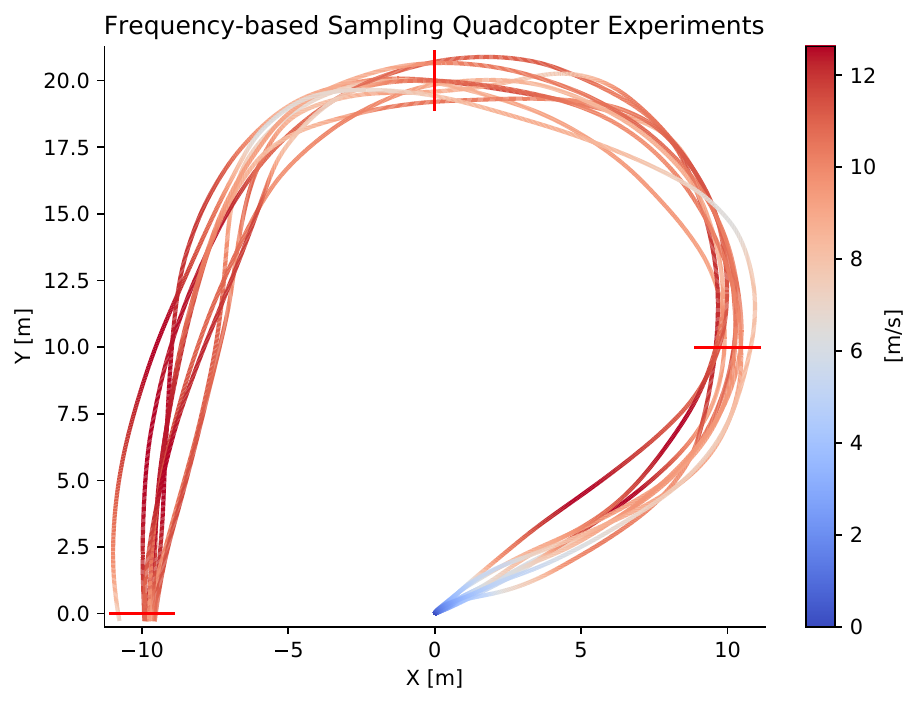}}
    \subfloat[Colored \label{fig:fastest_quad_colored_control}]{
        \includegraphics[width=0.5\textwidth]{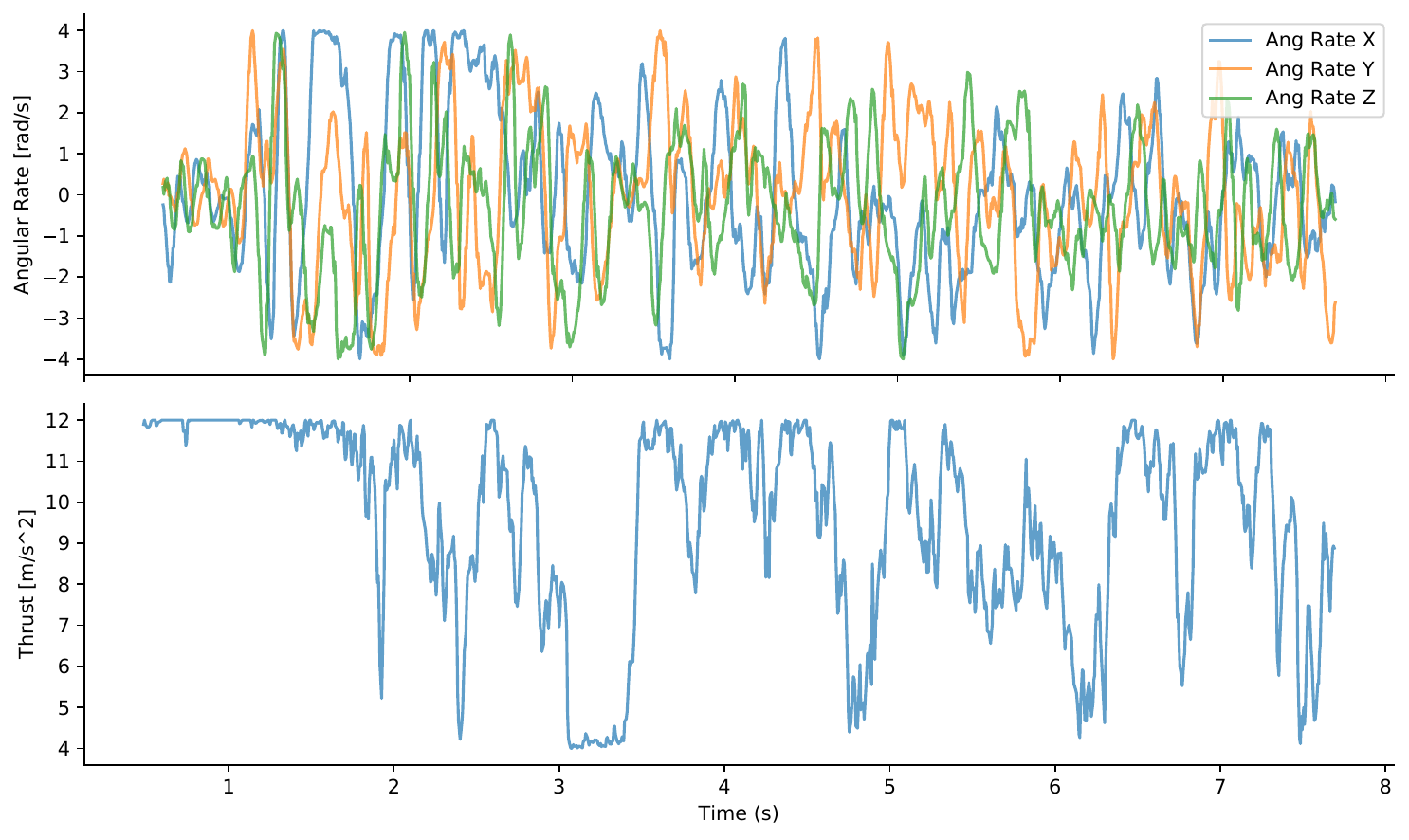}}
    \caption{Quadrotor flights using both types of sampling. Each controller is showing the first 9 successful attempts flying through this course. The red bars indicate gate positions that the quadrotor was expected to fly through. The left side graphs show the trajectories taken by the two controllers, colored according to the ground velocity of the quadrotor. On the right side, the controls that achieved the fastest lap for each controller are shown. Note that the throttle signal is much smoother for the Colored sampling when compared to Gaussian.
    }
\end{figure*}

\subsection{Simulated Quadrotor Results}
\label{subsec:sim_quad_results}
We conducted additional experiments on a quadrotor platform in simulation.
The simulation environment used is the Flightmare simulator \cite{song2021flightmare}, shown in \cref{fig:hw_vehicle}.
The dynamics function $\vb{F}\left(\vb{x}_t, \vb{u}_t\right)$ used within \ac{MPPI} had a 13-dimensional state of $\{p_x, p_y, p_z, v_x, v_y, v_z, q_w, q_x, q_y, q_z, \omega_x, \omega_y, \omega_z \}$, where $p$ is position [\si{\m}], $v$ is linear velocity [\si{\m\per\s}], $q$ is a quaternion representation of orientation, and $\omega$ are angular rates [\si{\radian\per\second}].
It was assumed a low-level controller existed on the angular rates that could achieve a first order response of $\tau = 0.25$, and the control space was angular rates [\si{\radian\per\second}] and thrust [$\si{\meter\per\second^2}$].
\changed{Both MPPI algorithms were run with $I = 2$ iterations, $M = 4096$ samples, $\lambda = 0.3$, $dt = 0.01\si{\s}$, and $T = 150$.}
The quadrotor was to fly through gates known a-priori.
The cost function was a combination of factors such as distance to next gate, maintaining height of $2 \si{\m}$ above the ground, tracking a desired velocity of $9 \si{\m\per\s}$, minimizing the deviation of the thrust axis from vertical to maintain stable flight, and crash costs for hitting the gate,
\begin{equation}
\begin{split}
    \label{eq:quad_cost_function}
    q(\vb{x}) &= a_1 \text{Heading}(\vb{x}) + a_2 \text{Height}(\vb{x}) + a_3 \text{Speed}(\vb{x})\\
    &+ a_4 \text{Stabilize}(\vb{x}) + a_5 \text{GateCrash}(\vb{x}) \\
    &+ a_6 \text{Waypoint}(\vb{x}) + a_7 \text{Path}(\vb{x}).
\end{split}
\end{equation}

When running \ac{MPPI} with Gaussian sampling, we used standard deviations of $[0.3, 0.3, 0.3, 3.5]$ for the angular rates and thrust respectively.
\changed{Note that the high thrust standard deviation is required for the quadrotor to adjust quickly when performing turns and reach quick lap times.}
As seen in \cref{fig:quad_gaussian_results}, the quadrotor successfully flies through the gates.
Looking at the controls in \cref{fig:fastest_quad_gaussian_control}, we see that the controls end up being quite chattery, especially in thrust.
This sort of chatter may lie outside of the control bandwidth of the quadrotor (i.e. going from $5 \si{\m\per\s^2}$ to $11 \si{\m\per\s^2}$ in $0.02 \si{\s}$ is not possible given the simulated motors) and can cause wear to physical motors over time.

\begin{figure*}[htbp!]
    \centering




    \includegraphics[width=0.28\linewidth,clip,trim=5 5 510 280]{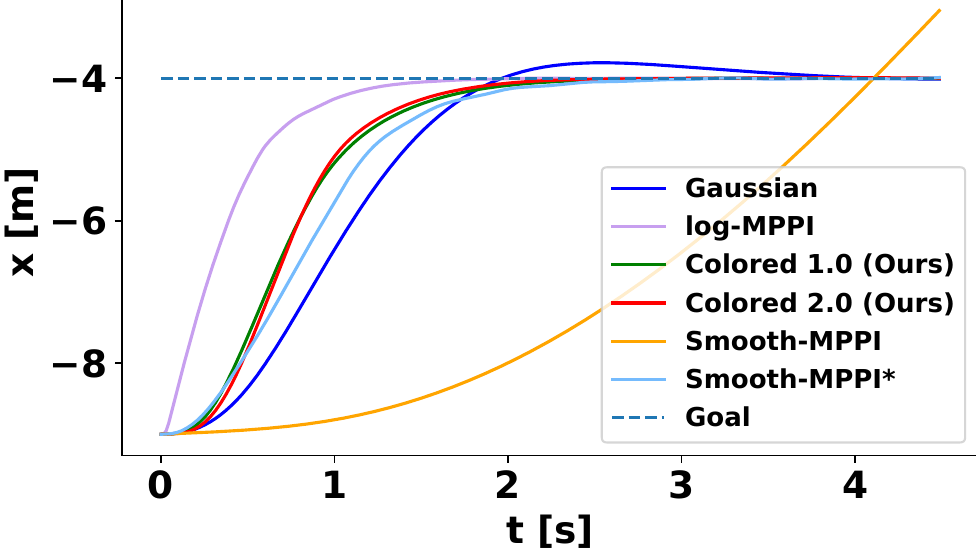}
    \includegraphics[width=0.3\linewidth,clip,trim=495 5 0 285]{Figures/Double_Integrator/double_integrator_std_dev_1.5_all_methods.pdf}
    \includegraphics[width=0.3\linewidth,clip,trim=5 5 0 285]{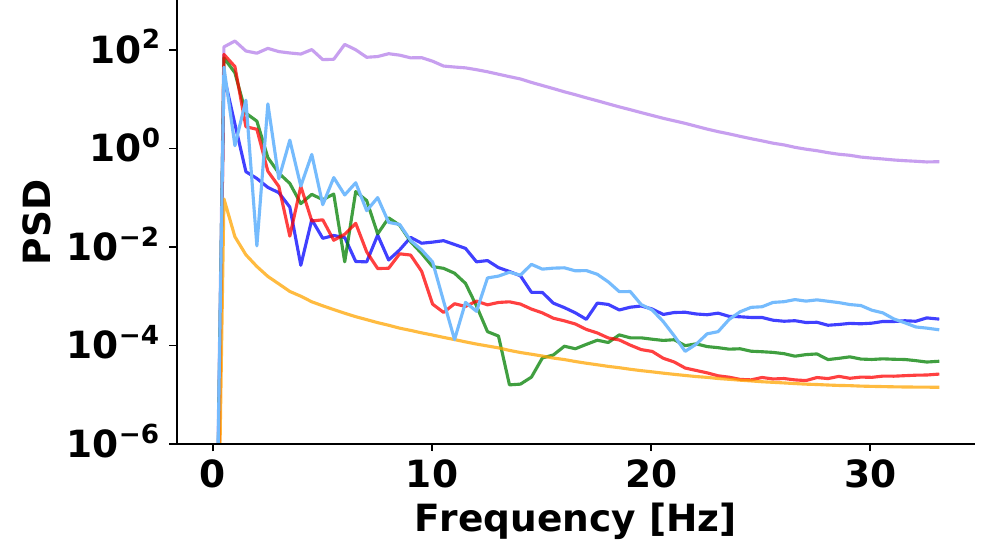}

    \caption{\changed{State trajectories \textbf{(Left)}, control trajectories \textbf{(Middle)}, and \ac{PSD} \textbf{(Right)} from the double integrator system with standard deviation set to $1.5$. The \ac{PSD} is shown for the first 2 seconds of the control trajectories to highlight the frequencies used to reach steady-state. Log-\ac{MPPI} reaches the goal fastest but does so with lots of high frequency controls. The Colored methods using $\gamma = 1.0$ and $2.0$ is able to react much quicker compared to Gaussian sampling by producing a much larger initial control spike and subsequent control dip to slow down to reach the goal without overshooting. Smooth-\ac{MPPI}* does reach the goal sooner than Gaussian sampling but has larger control deviations around steady state. Finally, the Smooth-\ac{MPPI} controller does not perform well with $\sigma = 1.5$, keeping the control pretty close to $0$.  The PSD shows that Colored control trajectories have lower contributions from high frequency components compared to Gaussian, Log-\ac{MPPI}, and Smooth-\ac{MPPI}*.}}
    \label{fig:double_integrator}
\end{figure*}

When running \ac{MPPI} with \changed{Colored} sampling, we used standard deviations of $[0.2, 0.2, 0.2, 0.5]$ and exponents $\gamma$ of $[0.01, 0.01, 0.01, 0.5]$ for the angular rates and thrust respectively based on best observed performance.
\changed{Since the quadrotor is more reactive to control inputs than the hardware ground vehicle, lower exponents allow us to make sharp maneuvers while still reducing control chatter.}
While the lap times do not show much difference, we can see a difference in \cref{fig:fastest_quad_colored_control} of the controls of the fastest lap.
Unlike the throttle of the Gaussian sampling case, we have lower frequency controls in throttle even when choosing a relatively small exponent of $0.5$.
These are more likely to be within the control bandwidth of the motors and we only see a $0.15\si{\second}$ average time~loss from \changed{Colored} sampling in \cref{tab:experiment_summary}.
The angular rates of Gaussian and \changed{Colored} sampling do not see much difference but this is in line with the angular rates exponent, $\gamma$ being set to $0.01$.
Looking at \cref{fig:quad_colored_results}, we can also see that the quadrotor achieved speeds of up to $12$ \si{\m\per\s} as it went towards the final gate compared to $10$ \si{\meter\per\second} from the Gaussian laps.

Looking at \cref{tab:experiment_summary}, we can see that \changed{Colored} sampling had the fastest time through the course but could not reliably achieve that time. The controllers were both run until they achieved 9 valid attempts. We see that the Gaussian sampling took many more attempts to get 9 valid laps than the \changed{Colored} sampling.
Overall, the quadrotor system sees little performance hit for smoother control inputs.

\subsection{Double Integrator Results}
\label{subsec:double_integrator}
Finally, we conduct a test on a simple double integrator system.
\changed{We tested six different controllers: \ac{MPPI} with a Gaussian distribution, log-\ac{MPPI} from \cite{mohamed2022autonomous}, Smooth-\ac{MPPI} from \cite{kim2022smooth} as well as a modification described below, and \ac{MPPI} with Colored sampling and colored exponents $\gamma = 1.0$ and $2.0$.
The \ac{MPPI} controllers had $M = 4096$ samples, $I = 1$ iteration, $dt = 0.015$\si{\s}, $T = 65$, and $\lambda = 1$ for all sampling techniques.}
The systems started at $-9$ \si{\m} and had a quadratic state cost of
\begin{align}
    q(\vb{x}) &= 5\PP{\vb{x_0} + 4}^2 + 0.5\vb{x_1}^2.
\end{align}

\changed{We conducted comparisons of all methods at $\sigma = 0.5$, $1.5$, and $3.0$ to see how adjustments to variance affect the performance of each controller.
We reran all the controllers $1000$ times to also see the variance caused by Monte-Carlo sampling which is reported in \cref{tab:double_integrator}.
Note that for Smooth-\ac{MPPI}, we saw poor performance using these standard deviations as the control would barely deviate from $0$.
We found that by dividing the provided standard deviation by $dt$, $\sigma^* = \frac{\sigma}{dt}$, performance fell in line with the other methods; we denote this change to the original method as Smooth-\ac{MPPI}* and include both in our comparisons.
In \cref{fig:double_integrator}, we visualize the state and control trajectories that achieved the minimum cost for controllers using $\sigma = 1.5$.
In addition, we also show the \ac{PSD} of the first 2 seconds of each control trajectory to visualize how much each control frequency contributes to the trajectory as it reaches the goal state.
At this $\sigma = 1.5$, log-\ac{MPPI} reaches the goal first but uses extremely large controls that would be impossible  to achieve on hardware systems.
Log-\ac{MPPI} also is extremely jerky in attempting to slow down, as evidenced by high power concentration for large frequencies in the \ac{PSD}.
The Colored sampling distributions with exponents $\gamma = 1$ and $2$ reach the goal next with smaller peak controls than log-\ac{MPPI}.
We see that the distribution with $\gamma = 2$ reaches the goal slightly faster and has lower power concentraion at frequencies above $20 \si{\Hz}$.
They both have smaller control deviations from $0$ at steady state compared to Smooth-\ac{MPPI}* and log-\ac{MPPI}.
Smooth-\ac{MPPI}* is the next to reach steady state but has oscillatory control behavior at steady state that is larger than all the other methods.
Gaussian sampling ends up overshooting the goal and the control trajectory is not as reactive as log-\ac{MPPI} or the Colored distributions.
Finally, Smooth-\ac{MPPI} as described in \cite{kim2022smooth} barely reaches the goal in the time allotted.
The control trajectory does not deviate from $0$ by very much and has low power over all frequencies, showing that it is unable to respond quickly with this sampling standard deviation.
Overall, Colored sampling allows the controller to be more reactive compared to Gaussian sampling while also being less oscillatory than both log-\ac{MPPI} and Smooth-\ac{MPPI}* at the same standard deviation.

We show the results across standard deviations of $0.5$, $1.5$, and $3.0$ in \cref{tab:double_integrator}.
The metric used in \cref{tab:double_integrator} for comparing the algorithms at various standard deviations is the accumulated cost of the states reached and we report the standard deviation of these costs over 1000 attempts.
Depending on the standard deviation, log-\ac{MPPI}, smooth-\ac{MPPI}, and smooth-\ac{MPPI}* can have orders of magnitude more variance in cost when tested multiple times compared to both Gaussian and Colored sampling.
Increasing the control standard deviation improves all methods in general but the best controller changes as well.
Our Colored sampling distributions maintain good relative performance, only losing to log-\ac{MPPI} at higher standard deviations.
In addition, we see in \cref{fig:double_integrator} that our method does not rely on control impulses or exhibit oscillatory behavior like controllers that do minimize the accumulated cost better.
Interestingly, Smooth-\ac{MPPI} becomes less consistent as the standard deviation increases but seems to have the best performance on average around $\sigma = 1.5$.
Looking at the Smooth-\ac{MPPI}* results and \cref{fig:double_integrator}, this is most likely due to the standard deviation choices being too small to properly excite the double integrator system.
Finally, there is some computational overhead to running Colored sampling compared to other methods, but the increase in computation time at around $0.12 \si{\milli\s}$ seems to fall within the RTX 3080's performance variance and is only noticeable when the dynamics and cost function are so simple.
While further tuning for each controller is possible to address any specific issue, we see that our proposed method performs well across a variety of standard deviations, both in terms of cost and control trajectory shape.}
\begin{table}[ht!]
\centering
\caption{\changed{Double Integrator Results}}
\label{tab:double_integrator}
\footnotesize
\sisetup{round-mode=places,separate-uncertainty,
detect-weight=true
}
\begin{tabular}{
c
c
S[round-mode=uncertainty,round-precision=3,round-pad=false,tight-spacing=true]
S[round-mode=uncertainty,round-precision=2,round-pad=false]
}
\toprule
\text{Std. Dev.} & \text{Sampling} & \text{Accumulated Cost} & \text{Calc. Time (ms)}  \\ \midrule
\multirow{6}{*}{0.5} & Gaussian &  \num{27919 \pm 33.1048} & \num{0.259161 \pm 0.225} \\
                     & Smooth-MPPI & \num{61328.7 \pm 1766.76}\text{0} & \num{0.295898 \pm 0.165282} \\
                     & Smooth-MPPI* & \num{21853.6 \pm 19189.7}\text{00} &  \num{0.294591 \pm 0.231628} \\
                     & Log-MPPI & \num{24545.3 \pm 40.2946} &  \num{0.297113 \pm 0.12421} \\
                     & \bfseries Colored 1.0 & \bfseries \num{13569.7 \pm 65.3729} & \bfseries \num{0.378722 \pm 0.156425} \\
                     & Colored 2.0 & \num{14201.7 \pm 61.2216} & \num{0.378722 \pm 0.156425} \\
\midrule
\multirow{6}{*}{1.5} & Gaussian & \num{13815.5 \pm 64.594} & \num{0.258975 \pm 0.18943} \\
                     & Smooth-MPPI &  \num{44401 \pm 3232.29}\text{0} & \num{0.292831 \pm 0.154399} \\
                     & Smooth-MPPI* & \num{11446.3 \pm 71.1} & \num{0.293511 \pm 0.17597} \\
                     & \bfseries Log-MPPI & \bfseries \num{7316.06 \pm 324.471} & \bfseries \num{0.298557 \pm 0.192854} \\
                     & Colored 1.0 & \num{10636.9 \pm 74.4388}  & \num{0.379149 \pm 0.177628} \\
                     & Colored 2.0 & \num{11081.2 \pm 75.5951} & \num{0.3791 \pm 0.220055} \\
\midrule
\multirow{6}{*}{3.0} & Gaussian & \num{10815.1 \pm 75.70} & \num{0.258632 \pm 0.21} \\
                     & Smooth-MPPI & \num{68088.2 \pm 7521.73}\text{0} & \num{0.29331 \pm 0.132446} \\
                     & Smooth-MPPI* & \num{10505.5 \pm 75.5529} & \num{0.293515 \pm 0.212292} \\
                     & \bfseries Log-MPPI & \bfseries \num{6963.46 \pm 380.457} & \bfseries \num{0.306633 \pm 0.155597} \\
                     & Colored 1.0 & \num{9191.69 \pm 69.6006} & \num{0.380484 \pm 0.331644} \\
                     & Colored 2.0 & \num{9699.99 \pm 70.19}  & \num{0.379612 \pm 0.249113}
\end{tabular}
\end{table}
\vspace{-0.3cm}
\section{Conclusions and Future Work}
\label{sec:conclusion}
In this work, we proposed use of a new sampling distribution with \ac{MPPI} to facilitate smoother controls and a larger coverage of the state space. This \changed{Colored} sampling distribution was shown to have nearly identical update laws when used in \ac{MPPI} as the Gaussian distribution.
This distribution \removed{is also} has a tunable amount of higher frequency components which provides it the ability to be used across various dynamical systems.
In our hardware results, our vehicle is unable to effectively maneuver when provided high-frequency samples from Gaussian distributions, reducing the reachable state space significantly.
Meanwhile, our \changed{Colored} sampling distribution demonstrated tight turns only achievable by sampling near the limits of the steering wheel for multiple timesteps in a row.
On the simulated quadrotor, we see that by lowering our sampling exponents, we can achieve nearly similar performance to Gaussian-based \ac{MPPI} while maintaining smoother throttle control.
Finally, in the double integrator experiments, \changed{Colored sampling is able to achieve more extreme control trajectories than the Gaussian-based controller and produce lower frequency control trajectories than other methods such as log-\ac{MPPI} and Smooth-\ac{MPPI}.}
In the future, we plan to apply this sampling distribution on more advanced forms of \ac{MPPI} such as Tube-MPPI and RMPPI as well as combine with other multi-hypothesis distributions such as \ac{GMM} and Stein variational distributions.
\section{Acknowledgements}
\changed{
This research was developed with funding from the Jet Propulsion Laboratory, \acl{DARPA}, and the Office of Naval Research.
The views, opinions and/or findings expressed are those of the author and should not be interpreted as representing the official views or policies of the Department of Defense or the U.S. Government. Approved for public release and unlimited distribution.
}

\balance
\bibliographystyle{IEEEtran}
\bibliography{references}
\end{document}

%% file: packages.tex
\usepackage[ruled,vlined]{algorithm2e}
\usepackage{amsfonts}       
\usepackage{amssymb}
\usepackage{balance}
\usepackage{booktabs} 
\usepackage{graphicx}
\usepackage{float}
\usepackage{mathtools}
\usepackage{mathrsfs}
\usepackage{multirow}
\usepackage[caption=false, font=footnotesize]{subfig}
\usepackage{siunitx}
\usepackage[normalem]{ulem}
\usepackage{usebib}
\usepackage[dvipsnames]{xcolor}

\iffalse
\usepackage[backref=section]{hyperref}
\else
\usepackage[backref=page]{hyperref}
\fi

\hypersetup{
    colorlinks=true,
    linkcolor=OliveGreen,
    filecolor=magenta,      
    urlcolor=blue,
    citecolor=purple,
}
\usepackage[capitalise]{cleveref}

\usepackage[nolist]{acronym}

\DeclareGraphicsExtensions{.png,.pdf}
\makeatletter

%% file: commands.tex
\newcommand{\vb}[1]{{\bf #1}}

\newcommand{\R}{\mathbb{R}}
\newcommand{\PP}[1]{\left(#1\right)}

\newcommand{\expf}[1]{\exp\left(#1\right)}
\newcommand{\Expectation}[2][]{\mathbb{E}_{#1}\left[#2\right]}

\DeclareMathOperator{\Shape}{\bf S}
\DeclareMathOperator{\J}{\bf J}
\DeclareMathOperator{\Jf}{\bf \check{J}}

\begin{acronym}
\acro{ADMM}{Alternating Direction Method of Multipliers}
\acro{AR}{AutoRally}
\acro{CEM}{Cross-Entropy Method}
\acro{CNN}{Convolutional Neural Network}
\acro{ConvLSTM}{Convolutional Long Short Term Memory}
\acro{CVaR}{Conditional Value-at-Risk}
\acro{DARPA}{Defense Advanced Research Projects Agency}
\acro{DDP}{Differential Dynamic Programming}
\acro{EKF}{Extended Kalman Filter}
\acro{EM}{Expectation Maximization}
\acro{ES}{Evolutionary Strategies}
\acro{GMM}{Gaussian Mixture Model}
\acro{GP}{Gaussian Process}
\acroplural{GP}[GPs]{Gaussian Processes}
\acro{GRU}{Gated Recurrent Unit}
\acro{FT}{Fourier Transform}
\acro{iCEM}{improved Cross-Entropy Method}
\acro{iFFT}{inverse Fast Fourier Transform}
\acro{IS}{Importance Sampler}
\acro{KL}{Kullback Leibler}
\acro{MOO}{Multi-Objective Optimization}
\acro{MPC}{Model Predictive Control}
\acro{MPPI}{Model Predictive Path Integral}
\acro{NLN}{Normal Log-Normal}
\acro{NN}{Neural Network}
\acro{PCA}{Principle Component Analysis}
\acro{PDF}{Probability Density Function}
\acro{PSD}{Power Spectral Density}
\acro{RCNN}{Region-based Convolutional Neural Network}
\acro{RL}{Reinforcement Learning}
\acro{RPN}{Region Proposal Network}
\acro{SQP}{Sequential Quadratic Programming}
\acro{VaR}{Value-at-Risk}
\acro{YOLO}{You Only Look Once}
\end{acronym}

\newcommand*{\pe}{\mathrel{+}=}
